\documentclass[journal]{IEEEtran}
\usepackage{amsmath,amsfonts}
\usepackage{algorithmic}
\usepackage{algorithm}
\usepackage{array}
\usepackage{textcomp}
\usepackage{stfloats}
\usepackage{url}
\usepackage{verbatim}
\usepackage{graphicx}
\usepackage{cite}

\usepackage{multirow}
\usepackage{subcaption}
\usepackage{bm}

\usepackage{lipsum}
\usepackage{times}
\usepackage{color}
\usepackage{amssymb}
\usepackage{makecell}
\usepackage{diagbox}
\usepackage{booktabs}
\usepackage[colorlinks,linkcolor=red]{hyperref}
\usepackage[table ]{ xcolor}
\usepackage{tablefootnote}
\usepackage{xspace}
\usepackage{enumitem}

\begin{document}

\title{CLIP-SCGI: Synthesized Caption-Guided Inversion for Person Re-Identification}

\author{Qianru Han, Xinwei He, Zhi Liu, Sannyuya Liu, Ying Zhang, Jinhai Xiang$^{\ddagger}$
\thanks{$^{\ddagger}$\ Corresponding author: Jinhai Xiang.}
\thanks{Q. Han,  X. He, Y. Zhang and J. Xiang are with the College of Informatics, Huazhong Agricultural University, Wuhan 430070, China. 
E-mail: qianruhan@webmail.hzau.edu.cn, \{xwhe, zy, jimmy\_xiang\}@mail.hzau.edu.cn.}

\thanks{Z. Liu and S. Liu are with the National Engineering Research Center of Educational Big Data, Central China Normal University, Wuhan 430079, China. 
E-mail:  zhiliu@mail.ccnu.edu.cn,  lsy.nercel@gmail.com.}

}



\maketitle

\begin{abstract}
Person re-identification (ReID) has recently benefited from large pretrained vision-language models such as Contrastive Language-Image Pre-Training (CLIP). 
However, the absence of concrete descriptions necessitates the use of implicit text embeddings, which demand complicated and inefficient training strategies. 
To address this issue, we first propose one straightforward solution by leveraging existing image captioning models to generate pseudo captions for person images, and thereby boost person re-identification with large vision language models.
Using models like the Large Language and Vision Assistant (LLAVA), we generate high-quality captions based on fixed templates that capture key semantic attributes such as gender, clothing, and age. 
By augmenting ReID training sets from uni-modality (image) to bi-modality (image and text), we introduce CLIP-SCGI, a simple yet effective framework that leverages synthesized captions to guide the learning of discriminative and robust representations. Built on CLIP, CLIP-SCGI fuses image and text embeddings through two modules to enhance the training process.
To address quality issues in generated captions, we introduce a caption-guided inversion module that captures semantic attributes from images by converting relevant visual information into pseudo-word tokens based on the descriptions. 
This approach helps the model better capture key information and focus on relevant regions. 
The extracted features are then utilized in a cross-modal fusion module, guiding the model to focus on regions semantically consistent with the caption, thereby facilitating the optimization of the visual encoder to extract discriminative and robust representations. 
Extensive experiments on four popular ReID benchmarks demonstrate that CLIP-SCGI outperforms the state-of-the-art by a significant margin.

\end{abstract}


\section{Introduction}
\IEEEPARstart{P}{erson} re-identification (Re-ID) involves identifying a person across images taken from different cameras. 
It is one of the critical components in intelligent surveillance systems for public safety and security~\cite{ye2021deep}. 
Typically, Re-ID is formulated as a person retrieval task: given a person image from a probe set, the goal is to learn discriminative representations that can correctly match the image of the same person in the gallery set. 
However, in practice, due to the presence of disturbing factors such as lighting, pose and occlusion~\cite{sarfraz2018posesensitive}\cite{OCCU}\cite{2018Mask}, it is an extremely challenging task to learn discriminative features.

\begin{figure}[t]
\begin{center}

\setlength{\abovecaptionskip}{0.8cm}
\subfloat[Existing CLIP-ReID method]{
    \centering 
    \includegraphics[width=1.0\linewidth]{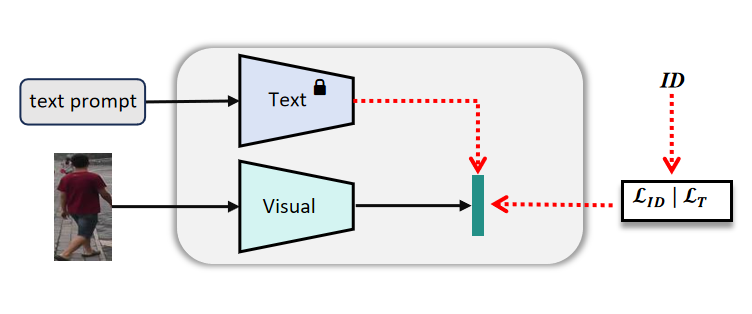}
    \label{fig_intro1}
}

\subfloat[Our CLIP-SCGI]{
    \centering
    \includegraphics[width=1.0\linewidth]{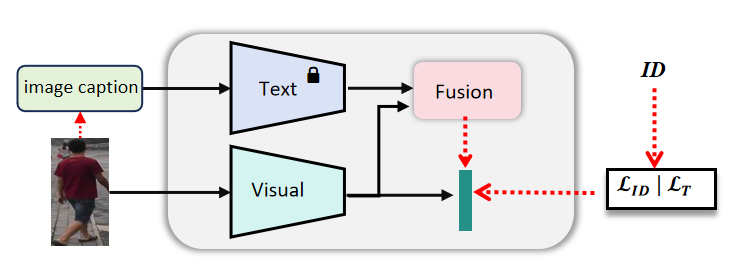}
    \label{fig_intro2}
}
\caption{Comparison between (a) the previous CLIP-ReID method, where the text prompt in the grey-highlighted area indicates learnable embeddings requiring pretraining first, and (b) our proposed CLIP-SCGI method, which combines explicitly generated text embeddings with image features for guidance.}
\label{fig_intro}

\vspace{-0.5cm} 
\end{center}
\end{figure}

Various strategies have been proposed to improve the representation learning process from both feature level and loss levels. At the feature level, global feature-based methods~\cite{global}\cite{sun2017svdnet} aim to learn global features to represent the person images, while the local-based methods~\cite{local1}~\cite{local2} aim to aggregate part-level local features. At the loss level, researchers have also presented various similarity loss functions for training. For instance, pairwise losses~\cite{pairwise} attempt to minimize the distance between positive pairs while maximizing the negative pair distances. Triplet loss~\cite{Triplet} enforces ranking order on the relative distances of anchor to positive and negative, respectively. 
Current methods often focus on improving single-modality feature learning and loss functions. 
Despite promising results, existing methods face the following limitations:
1) the semantic components of single-image modalities are inadequate for extracting comprehensive semantic information, potentially overlooking the inherent and fine-grained semantic attribute information across different samples, which restricts overall performance; 
2) ReID datasets are relatively small, and models trained on these datasets may easily overfit small irrelevant regions in the image, leading to poor generalizability to real yet diverse scenarios.



Recently, CLIP~\cite{clip} which is trained on 400 million image-text pairs with contrastive loss, has been demonstrated to produce strong generalizable representations for a wide array of downstream tasks.
Inspired by its success, CLIP-ReID~\cite{CLIP-ReID} explores its effectiveness in the ReID field and observes encouraging results. Nonetheless, most ReID datasets do not contain corresponding image-text pairs. 
To address this issue, they further leverage learnable prompts to describe the person images implicitly, and then present a two-stage training strategy to alternately fix the image and text encoders to optimize them
(see Fig.~\ref{fig_intro}\subref{fig_intro1}). Yet, such an implicit way to describe the image may lead to a compromise in performance and require a complicated and inefficient training process.

When performing ReID tasks, it is a natural human instinct to utilize textual descriptions to emphasize and distinguish the individuals in the images.
Indeed, regardless of the availability, concrete descriptions (or captions) may naturally be a better fit for supervising the network to learn discriminative and robust features for ReID than the single ID index and the implicit learnable text embeddings. First, it enables flexible and robust learning of ID-specific embeddings. With captions as the guidance, the model can now be explicitly enhanced by shared semantics across samples and thereby produce more diversified features. In contrast, representation learning based solely on single IDs makes it easily overfits the most pronounced attributes like clothes.   
Second, it helps the model to produce more expressive representations of the person image. Inherently, the captions can provide a more comprehensive understanding of the distinctive aspects of the persons, e.g., various visual attributes and contextual information, which assist the model in focusing on specific identity-related features, thereby overcoming challenges like variations in pose and lighting conditions.

Thanks to large vision language models, we can easily generate high-quality synthetic captions for each image to enrich a ReID dataset with only visual images. Any off-the-shelf pretrained image captioning model can be utilized. In this paper, we utilize LLAVA~\cite{llava} to generate the captions for each image. 
The \emph{data augmentation} process of converting a uni-modal (\emph{i.e.}, image) dataset to a bimodal (\emph{i.e.}, image and text) enables us to leverage state-of-the-art large vision and language models (\emph{e.g.}, CLIP) to learn robust representations with minimal effort. 
Note that we only have to generate these captions offline for our training data and use them to guide the representation learning, which is very flexible in practice and does not incur extra inference costs.

Based on the synthesized captions, we introduce a simple yet effective framework named CLIP-SCGI by exploiting \textbf{S}ynthesized \textbf{C}aption-\textbf{G}uided \textbf{I}nversion for person reID (Fig.~\ref{fig_intro}\subref{fig_intro2}). Our framework leverages the pretrained CLIP model, which has demonstrated superior capacity in deriving discriminative representations for ReID.
Given a person image and its corresponding synthesized text, we first feed them into the visual encoder and text encoder, respectively, to produce image and text embeddings. We then fused the two modalities to generate more accurate guidance. 
Despite the effectiveness of large vision language models, the generated descriptions can still contain errors. These inaccuracies in the textual descriptions can negatively impact the training process and overall model performance. Therefore, addressing the impact of incorrect captions is crucial.
To address this issue, we introduce a novel approach inspired by textual inversion~\cite{textinversion}~\cite{baldrati2023zero}. Our method first maps the person image into a pseudo-word $S_{*}$ guided by the synthesized captioning embeddings and then insert it with a fixed prompt ``a photo of [$S_{*}$] person''.  
Consequently, the corresponding concepts exhibited in the visual space are collected by the text embeddings to provide more accurate guidance. 
We further feed the prompt into the text encoder again to encode it into the text embedding space, which will serve as a query for fusion with the visual features. Based on the final multi-modal embeddings, we train the whole framework with the widely-used loss functions: cross-entropy loss and triplet loss.

Our framework has the following desirable properties. First, it opens the possibility of training more expressive and flexible representations for ReID with synthesized captions, which alleviates the overfitting issue to some extent compared with existing identity-based training.
Second, it is free from annotations and does not increase the inference time and costs since it merely utilizes the synthesized caption at the training phase to guide the feature extraction network. After training, this caption-guided branch can be removed safely. 
Lastly, our framework is simple yet effective. With synthesized caption-guided inversion, our model can better harvest the latent corresponding semantic cues in the visual image and assist in learning more discriminative representations for ReID task. Extensive experiments are conducted on existing popular benchmarks and prove the superiority of our methods. For instance, on dataset MSMT17, our framework achieved new state-of-the-art results, reaching 88.2\% in mAP.

In summary, our contributions are summarized as follows:
1) A simple and effective representation learning scheme, which is guided by synthesized captions, is proposed for ReID. It can derive more expressive and more flexible learning compared with existing methods. 
2) We design a simple and effective framework based on CLIP to fully harvest the correlations between visual and semantic features to enhance the model deriving discriminative presentations. Experiments on several popular datasets prove its effectiveness.

\section{Related work}

\subsection{Person Re-identification}

Person re-identification (Re-ID) has gained significant attention due to its applications in intelligent surveillance systems~\cite{ye2021deep}. Traditional CNN-based approaches have primarily focused on feature learning~\cite{8237611}\cite{8607050} and metric optimization ~\cite{li2017person}\cite{Martin2012Large}\cite{7298832} but often struggle with overfitting and irrelevant region highlighting.
To enhance feature representation, prior knowledge has been integrated into networks~\cite{PCB} to extract finer-grained features by combining global and local information from pedestrian images.
With the advent of the Transformer model, Re-ID has seen improved performance due to its self-attention mechanisms~\cite{TransReID}\cite{AAformer}\cite{li2022pyramidal}\cite{HAT}, proving effective in pedestrian tasks and often outperforming CNNs on large-scale public datasets. Notably, TransReID utilizes only Transformer architectures, achieving superior results compared to CNN models. AAformer introduces a Transformer-based framework for the automatic alignment of local regions in Re-ID. 
However, attention mechanisms in existing methods often fail to effectively capture discriminative features, particularly in the presence of background occlusions, and predominantly rely on single visual modalities, neglecting comprehensive multi-modal fusion.
CLIP-ReID~\cite{CLIP-ReID} was the first to leverage multi-modal information but primarily optimized image encoders without fully exploiting the textual semantic information embedded in the images. PromptSG~\cite{yang2024pedestrian} guides the model to focus on semantically relevant regions but falls short in capturing critical pedestrian details through simple pseudo token learning. 
In contrast, our approach explicitly extracts key person information, enhancing the model's attention and learning capabilities.

\subsection{Vision-Language learning}

The advent of vision-language models, particularly with CLIP, has significantly advanced visual representation learning by leveraging extensive image-text datasets. Recent developments in large language models (LLMs)~\cite{2020t5}\cite{gpt3} have further propelled the evolution of Vision-Language Models (VLMs), which integrate separately pre-trained LLMs and vision foundation models to interpret both images and text.
LLaVA~\cite{llava} exemplifies impressive multimodal capabilities with minimal additional training. Unlike existing methods that rely on LLM-generated large-scale pre-training datasets for text-based cross-modal person re-identification~\cite{APTM}\cite{UNIPT}~\cite{tan2024harnessing}, our approach focuses on employing captions specifically for image-based ReID. This methodology aims to exploit the semantic information embedded in images rather than depending solely on pre-existing datasets.
Research like CLIP-ReID\cite{CLIP-ReID} has shown that incorporating language modalities enhances visual concept learning and facilitates discriminative feature extraction. Our method distinguishes itself by enriching cross-modal information through sample-level and feature-level operations. We also introduce a novel text filtering technique to address inaccuracies in captions generated by LLMs, ensuring high accuracy in the synthesized captions used in our framework. This approach leverages the capabilities of the CLIP model and implements modal fusion techniques to integrate this enriched information, thereby enhancing the overall performance of image-based ReID.

\subsection{Representing image as one word}

Textual inversion was first introduced to guide personalized generation using a set of user-provided images~\cite{textinversion}. By learning pseudo-words in the text embedding space of a text-to-image generation model, this method captures key concepts for personalized output.
Beyond text-to-image generation, textual inversion has attracted attention in image retrieval, especially in zero-shot tasks where it is pre-trained on large, unlabeled image datasets~\cite{saito2023pic2word}\cite{baldrati2023zero}\cite{tang2023contexti2w}\cite{yang2024pedestrian} . In text-based person retrieval, methods like the lightweight Textual Inversion Network (TINet)~\cite{liu2024word4per} have been developed for cross-modal retrieval.
In image-based person re-identification, distinguishing the foreground from the background is key to reducing interference. Existing methods struggle with representing and integrating visual features effectively. Our approach uses captions to guide the generation of pseudo-words, which are incorporated into a feature fusion module that combines textual and visual data. This enhances the model's ability to learn a more discriminative feature space, improving both accuracy and robustness in person re-identification.

\section{Method}

\begin{figure}[tp]
\begin{center}
\includegraphics[width=3.4in]{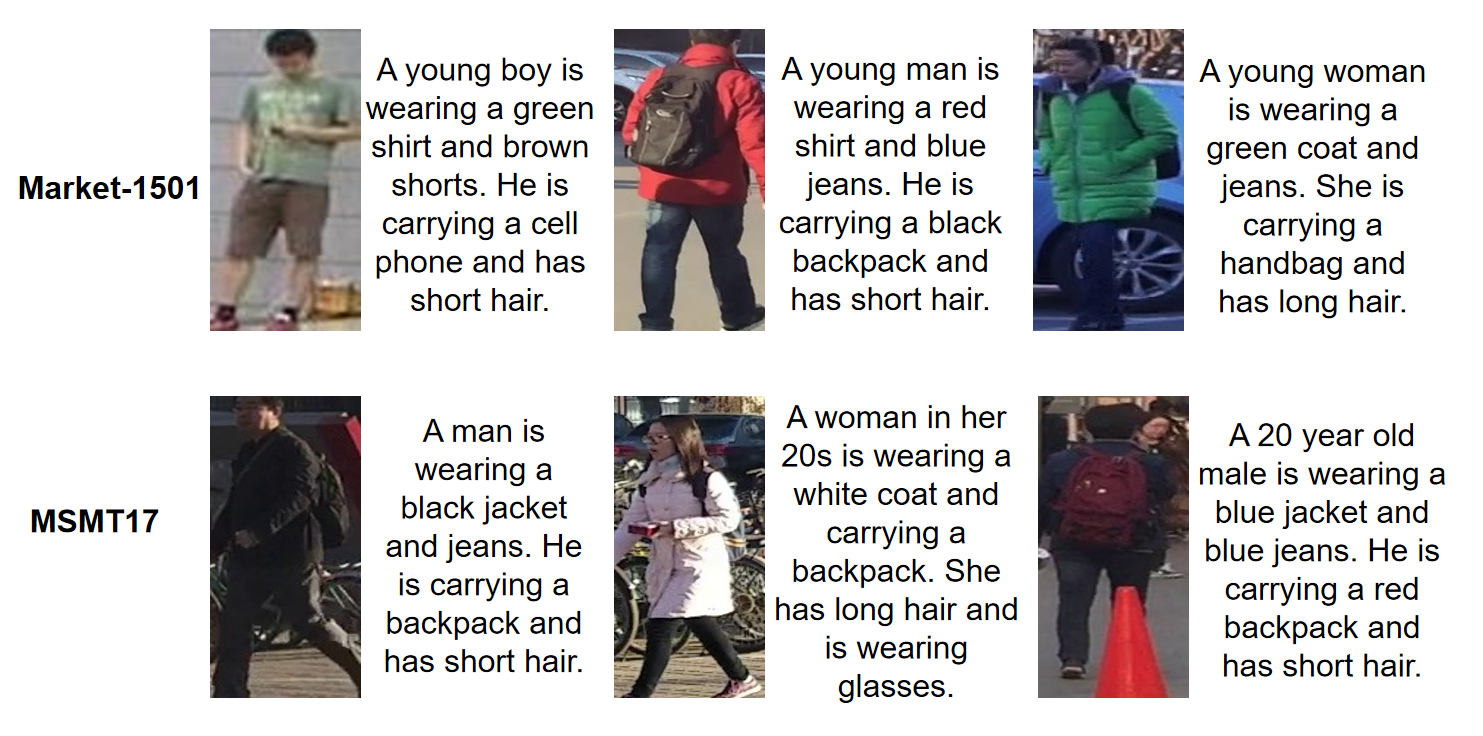}
\end{center}
\caption{Caption Generation results. The first row shows the captions generated for the Market1501 dataset, while the second row depicts the results for the MSMT17 dataset.}
\label{fg:caption}
\end{figure}

\begin{figure*}[htbp]
\begin{center}
\includegraphics[width=1\textwidth]{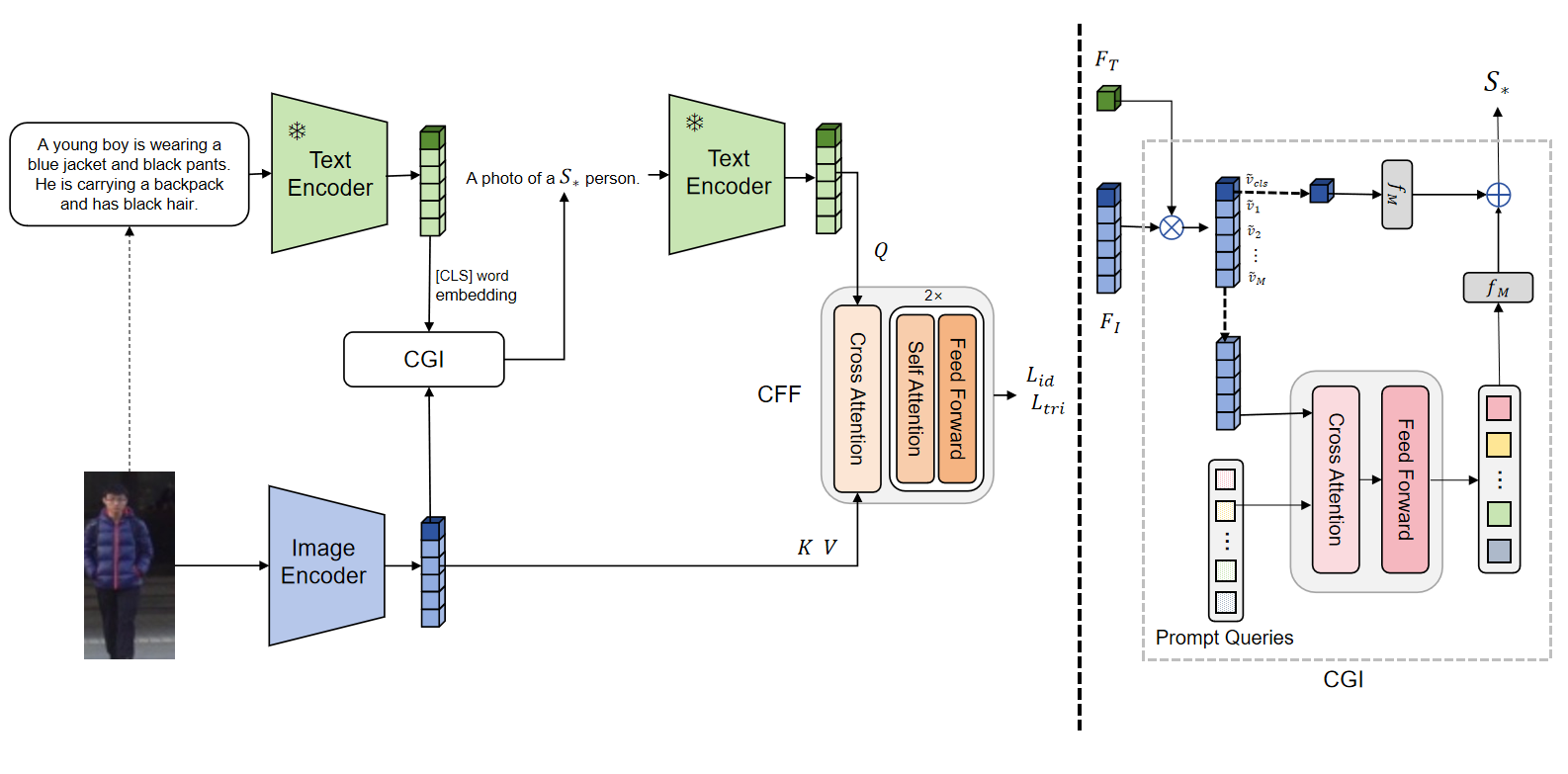}
\end{center}
\caption{Overview of CLIP-SCGI. It enhances CLIP with two core modules named caption-guided inversion(CGI) and contextual feature fusion(CFF) for ReID. The CGI module transforms the images to a pseudo-word under the guidance of captions, while the CFF module fuses the pseudo-word embeddings and images to facilitate learning a discriminative global representation.}
\label{fg:method}
\end{figure*}

In this section, we present a detailed description of our caption-guided framework for the person re-identification.
Fig.~\ref{fg:method} gives the overview of our framework. 
During training, our framework takes the two modalities as input: image and the corresponding caption pairs. We argue that caption embeddings capture the high-level semantic attributes of the image and provide more fine-grained supervision than existing works based on the category ID index. 
Yet, existing ReID datasets only contain person image data. 
To augment these unimodal datasets into bi-modal ones to enable the training, we turn to large pre-trained vision and language models to synthesize high-quality captions for each training image.  

\subsection{Synthesized Caption Generation}
The training process of our ReID framework requires image-text pairs.
However, most existing ReID benchmarks lack text modality, and annotating such datasets with textual information is prohibitively expensive.
To address this challenge, we leverage pretrained large multi-modal models to synthesize captions for each training image offline. This approach allows us to create a novel bi-modal training set specifically tailored for ReID tasks.
Any off-the-self image captioning model can be leveraged. 
In this work, we use the strong pretrained model LLAVA~\cite{llava} to extract the semantic attribute information contained in the person images. More importantly, it is a multimodal instruction-following tool which can generate the concept of interests related to the persons in the image. 
For example, as illustrated in Fig.~\ref{fg:caption}, LLAVA provides nuanced descriptions that capture the key aspects of the person’s appearance, which significantly contributes to the model's ability to learn discriminative features for ReID.

Let $\mathcal{T}_\text{tr} = \{I_k\}_{k=1}^{K}$ denote the ReID training set, then for each image $I_k$ we feed it into LLAVA to generate the corresponding caption $T_k$. We perform image captioning for the whole training set in an offline manner for efficiency, resulting in an augmented multi-modal dataset $\hat{\mathcal{T}}_\text{tr} = \{(I_k, T_k)\}_{k=1}^{K}$. 
For ReID, person-related fine-grained attributes such as top, pants, shoes, and accessories are essential. Therefore, we input the instruction to LLAVA by specifying the character information while ignoring the irrelevant background.
For our experiments, the instruction used for captioning is:
 ``\emph{Describe the most obvious one people using this template for no more than 45 words: A [age] [gender] is wearing [upper clothes].  [gender] is wearing [upper clothes] and [lower clothes]. The [gender] is wearing [shoes] and carrying a [others]. The [gender] has [hair length] [and whether wear glasses].}''
 With the augmented training set $\hat{\mathcal{T}}_\text{tr}$, we then train our framework CLIP-SCGI under the guidance of synthesized captions.

\subsection{CLIP-SCGI}

\subsubsection{Overview} Fig.~\ref{fg:method} illustrates the details of our framework. As shown, CLIP-SCGI consists of three main components: CLIP, CGI (Caption Guided Inversion module), and CFF (Contextual Feature Fusion module). CLIP is used for embedding the image and text into an aligned feature space, while CGI and CFF support multi-modal training, guiding our model to pay attention to the fine-grained semantic information. Below we give the details of each component. 

\subsubsection{CLIP} Similar to CLIP-ReID, our framework leverages large pretrained vision and language model CLIP as the backbone.
In essence, CLIP $f(\cdot)$ consists of an image encoder $f_{img}(\cdot)$ and a text encoder $f_{txt}(\cdot)$. Given an image-text pair $(I_k, T_k) \in \hat{\mathcal{T}}$, we feed the image and text into CLIP and obtain their corresponding embeddings:
\begin{equation}
    F_{img}^k = f_{img}(I_k), \quad F_{txt}^k = f_{txt}(T_k)
\end{equation}
where $F_{img}^k$ = \{\(v_{cls}\), \(v_{1}\), \(v_{2}\), ..., \(v_{M}\)\} represents a sequence of visual tokens for the input image $I_k$ with \(v_{cls} \) being the global one and $v_{i}$ being local one, $F_{txt}^k = \{t_{sos}, t_{1}, t_{2}, ..., t_{eos}\}$ denotes the sequence of corresponding text token embeddings which includes the extra padded special [SOS] and [EOS] tokens.  CLIP models have trained on large-scale image and text data with contrastive loss, therefore the learned embeddings between text and image are naturally aligned in the same feature space. 


\subsubsection{Caption Guided Inversion Module}
After obtaining the text and image embeddings with pretrained CLIP models, we then design a caption-guided inversion module that distills the semantic features in the person image according to the synthesized captions, which will be used to train the visual encoder for discriminative representations for ReID. 
This strategy is inspired by textual inversion~\cite{textinversion}.
In essence, the objective of textual inversion is to learn an inversion mapping $f_\theta(\cdot)$ from a point in the CLIP image space \(\mathcal{I}\) to a point in its corresponding word embedding input space \(\mathcal{W}\): 
\begin{equation}
f_\theta : {z\in \mathcal{I} } \to \mathcal{W} 
\end{equation}
Here, \(\mathcal{I}\) denotes the CLIP embedded image space, \(\mathcal{T}\) the CLIP embedded textual space,
\(z_k=h^{I}(I_k)\) is the embedding of an image \(I_k\) into \(\mathcal{I}\) , and similarly \(h^{T}(S)\) is the embedding of the caption $S$ into \(\mathcal{T}\). \(\mathcal{W}\) denotes the space used to embed input word tokens into CLIP. 
However, the inversion process may result in a pseudo word be disturbed by a lot of irrelevant factors such as illuminations, background and weather in the image, which may limit the performance.  

To address this issue, we propose textual inversion guided by the synthesized captions. Specifically, we first refine the image features by utilizing a point-wise multiplication operation between image and caption embeddings:
\begin{equation}
    \tilde{v} = v \odot t
\end{equation}
where \(\odot\) represents element-wise multiplication. 
This operation aims to adjust the weight of image features based on the information conveyed in the accompanying text, thereby directing visual attention towards the relevant content described in the text. Besides, the model can effectively alter the importance or emphasis of different visual features according to the textual context. 

Then we map the reweighed image features into a pseudo-word $S^*$ guided by the synthesized captioning embeddings. The generation of  $S^*$ considers both global and local features. The global feature \(\tilde{v}_{cls}\) is used to obtain global pseudo-word embeddings as tokens \(S_{global}^* = f_{M_g}(\tilde{v}_{cls})\), while the local features 
$\{\tilde{v}_{1}, \tilde{v}_{2}, \dots,\tilde{v}_{M}\}$ is used to obtain local pseudo-word embeddings as tokens $S_{local}^*$. We condition the prompt generation on visual local features \(\tilde{v}_{i}\)(i=1, 2, ..., M) through transformers layers, as shown in Fig.~\ref{fg:method}. Formally, we randomly initialize K learnable prompt queries \(Q=\{ q_{1}, q_{2}, ..., q_{K} \} \in \mathbb{R}^{K{\times} d}  \). Then these queries are fed into the transformers blocks, the queries served as query(Q), and the local image representations are served as key(K) and value(V), which can be formulated as follows:
\begin{equation}
    Q_C = Q + \text{Cross\text{-}Attn}(\text{LN}(Q, \tilde{v_{i}}, \tilde{v_{i}}))
\end{equation}
\begin{equation}
    P = Q_C +\text{FFN}(Q_C)
\end{equation}
where \(P=\{ p_{1}, p_{2}, ..., p_{K} \} \in \mathbb{R}^{K{\times} d}   \) is the generated prompts, Cross-Attention($\cdot$) denote the cross-attention operation, respectively. LN($\cdot$) is the Layer Normalization and FFN($\cdot$) is the fully connected network. The learnable Q can be updated during training through gradient back-propagation. 

The local pseudo-word embeddings are obtained by tokens \(S_{local}^* = f_{M_l}(Avg(P))\). These local pseudo-word embeddings are then combined with the global pseudo-word embeddings to form the pseudo-word \(S^* = S_{global}^* + S_{local}^*\). \(f_{M_g}\) and \(f_{M_l}\) is a three-layered fully-connected network. Following this, it is concatenated with the token embeddings of the prompt sentence, "a photo of a person", thereby forming the augmented prompt sequence \(\hat{S}^*\)=" a photo of a \(S^*\) person". Subsequently, \(\hat{S}^*\) is passed through the text encoder to generate textual inversion embeddings \(\{\hat{t}_{sos}, \hat{t}_{1}, \hat{t}_{2}, ..., \hat{t}_{eos} \}\), which can represent an input image embedding. we propose to minimize the contrastive loss with respect to the mapping network.
\begin{equation}
    \mathcal L_{con}= \mathcal L_{t2i} + \mathcal L_{i2t}
\end{equation}
\begin{equation}
    \mathcal L_{t2i}(T, I) = -\frac{1}{P} \sum_{i\in P}^{} log\frac{exp(s(T_i, I_p) )}{ {\textstyle \sum_{k=1}^{B}}exp(s(T_i, I_k))}
\end{equation}
\begin{equation}
    \mathcal L_{i2t}(I, T) = -\frac{1}{P} \sum_{i\in P}^{} log\frac{exp(s(I_i, T_p) )}{ {\textstyle \sum_{k=1}^{B}}exp(s(I_i, T_k))}
\end{equation}

where $P$ represents the set of positive samples, \(s(T_i, I_i) = T_i \cdot  I_i\)  refers to the similarity function that evaluates the distance between text and image.

\subsubsection{Contextual Feature Fusion}
To enable comprehensive interaction between image and text modalities, we introduce a multi-model fusion module~\cite{dou2022empirical}. The multi-model interaction encoder consists of a multi-head cross-attention layer and 2-layer transformer blocks.                                                                                      
The text representation \(\{\hat{t}_{sos}, \hat{t}_{1}, \hat{t}_{2}, ..., \hat{t}_{eos} \}\) served as query(Q), and the image representation \(v_{cls} \) are served as key(K) and value(V). The full interaction between image and text representations can be achieved by:

\begin{equation}
    h = \text{Transformer}(\text{Cross\text{-}Attn}(\text{LN}(Q, K, V)))
\end{equation}
where $h$ is the fused image and text contextual representations, LN($\cdot$) denotes Layer Normalization, and $\text{Cross-Attn}(\cdot)$ is the multi-head cross attention and can be realized by:
\begin{equation}
    \text{Cross\text{-}Attn}(Q, K, V) = softmax(\frac{QK^{T}}{\sqrt{d} })V
\end{equation}
where $d$ is the embedding dimension of tokens. 
After the fusion of image and text contextual representations, we employ the resulting features to compute cross-entropy loss.

\subsection{Optimization and Inference}

We employ the identity classification loss \(\mathcal L_{id}\) and triplet loss \(\mathcal L_{tri}\) ~\cite{he2020fastreid} for optimization: 
\begin{equation}
    \mathcal L_{id} = \sum_{k=1}^{N}-q_klog(p_k)
\end{equation}
\begin{equation}
    \mathcal L_{tri} = max(d_p-d_n+\alpha, 0)
\end{equation}
Here, $p$ represents the logits distribution obtained by the classifier, $q$ represents the ground truth distribution, and N is the total number of instances. 
During the inference process, for efficiency reasons, our captioning method only utilizes the phrase "a photo of a person" without employing additional captioning techniques to generate captions for the images. Therefore, the caption-guided branch can be removed safely without incurring extra computation costs during inference. The total loss to train our framework is given by the following equation:
\begin{equation}
    \mathcal L = \mathcal L_{id} + \mathcal L_{tri} + \mathcal L_{con}
\end{equation}

\section{Experiment}

\label{exp}

\setlength{\tabcolsep}{3.5mm}
\begin{table*}
\renewcommand\arraystretch{1.5}
\begin{center}
\caption{Comparison with the state-of-the-art CNN- and ViT-based models on Market-1501, MSMT17, DukeMTMC, and OccDuke datasets. The superscript star* means that the input image is resized to a resolution larger than 256x128. Note that all data listed here are without re-ranking.}
\label{tab:performance}
\begin{tabular}{l|l|l||c|c|c|c|c|c|c|c}
\specialrule{0.1em}{0pt}{0pt}  

\multirow{2}{*}{Backbone} & \multirow{2}{*}{Method} & \multirow{2}{*}{Reference} & \multicolumn{2}{c|}{MSMT17} & \multicolumn{2}{c|}{Market-1501} & \multicolumn{2}{c|}{DukeMTMC-reID} & \multicolumn{2}{c}{Occluded-Duke} \\ \cline{4-11}
 &  &  & mAP & top-1 & mAP & top-1 & mAP & top-1 & mAP & top-1 \\
\specialrule{0.1em}{0pt}{0pt}

\multirow{15}{*}{CNN}
 & OSNeT*\cite{osnet}& ICCV 2019 & 52.9 & 78.7 & 84.9 & 94.8 & 73.5 & 88.6 & - & - \\
 & Auto-ReID*\cite{AutoREID}& ICCV 2019& 52.5 & 78.2 & 85.1& 94.5 & - & - & - & - \\
 & HOReID\cite{HOReID}& CVPR 2020& - & - & 84.9 & 94.2 & 75.6 & 86.9 & 43.8 & 55.1 \\
 & ISP \cite{isp}& ECCV 2021& - & - & 88.6 & 95.3 & 80.0 & 89.6 & 52.3 & 62.8 \\
 & CDNet \cite{cdnet}& CVPR 2021& 54.7 & 78.9 & 86.0 & 95.1 & 76.8 & 88.6 & - & - \\
 & PAT \cite{PAT}& CVPR 2021& - & - & 88.0 & 95.4 & 78.2 & 88.8 & \textbf{53.6} & 64.5\\
 & CAL*\cite{cal}& ICCV 2021& 56.2 & 79.5 & 87.0 & 94.5 & 76.4 & 87.2 & - & - \\
 & CBDB-Net*\cite{CBDB-Net}& TCSVT 2021& - & - & 85.0 & 94.4 & 74.3 & 87.7 & 38.9 & 50.9 \\ 
 & ALDER*\cite{ALDER}& TIP 2021& 59.1 & 82.5 & 88.9 & 95.6 & 78.9 & 89.9 & - & - \\
 & LTReID*\cite{LTReID}& TMM 2022& 59.1& 82.5& 88.9 & 95.6& 78.9 & \textbf{90.5}& -&-\\
 & DRL-Net\cite{DRL-Net}& TMM 2022& 55.3 & 78.4 & 86.9 & 94.7 & 76.6 & 88.1 & 50.8 & \textbf{65.0}\\ 
 \cline{2-11} 
 & baseline &  & 60.7 & 82.1 & 88.1 & 94.7 & 79.3 & 88.6 & 47.4 & 54.2 \\
 & CLIP-ReID \cite {CLIP-ReID}&  AAAI 2023& 63.0& 84.4& 89.8& 95.7 & \textbf{80.7}& 90.0 & 53.5 & 61.0 \\
 & PromptSG\cite{yang2024pedestrian} &  CVPR 2024& 68.5& 86.0& 91.8& \textbf{96.6} & 80.4& 90.2 & - & - \\
 & Ours& —— & \textbf{80.9}& \textbf{87.9}& \textbf{92.1}& 95.7&  77.7 & 89.0& 47.4&60.4\\ \hline
\specialrule{0.1em}{0pt}{0pt}

\multirow{8}{*}{ViT} 
 & TransReID\cite {TransReID}& ICCV 2021&  67.4 & 85.3 & 88.9 & 95.2 & 82.0 & 90.7 & 59.2 & 66.4 \\
 & AAformer*\cite {AAformer}& arxiv 2021& 63.2 & 83.6 & 87.7 & 95.4 & 80.0 & 90.1 & 58.2 & 67.0 \\
 & DCAL\cite{DCAL}& CVPR 2022& 64.0 & 83.1 & 87.5 & 94.7 & 80.1 & 89.0 & - & - \\
 & PHA \cite{PHA}& CVPR 2023& 68.9& 86.1& 90.2& 96.1& -& -& -&-\\
 \cline{2-11}
 & baseline & & 66.1 & 84.4 & 86.4 & 93.3 & 80.0 & 88.8 & 53.5 & 60.8 \\
 & CLIP-ReID\cite {CLIP-ReID}& AAAI 2023& 73.4& 88.7& 89.6& 95.5& \textbf{82.5}& 90.0 &  \textbf{59.5}&  67.1\\
 & PromptSG\cite{yang2024pedestrian}& CVPR 2024& 87.2& 92.6& 94.6& 97.0& 81.6& 91.0 &  - &  -\\
 & Ours & —— &\textbf{88.2}& \textbf{92.9}& \textbf{96.0}& \textbf{97.6}& 81.6&  \textbf{91.3}&  59.2&  \textbf{67.5}\\ \hline
\specialrule{0.1em}{0pt}{0pt}  
\end{tabular}
\end{center}
\end{table*}

\subsection{Experiment Setup}
\subsubsection{Datasets and Evaluation Protocol}
We evaluate our method on four popular person re-identification datasets, including Market-1501~\cite{market1501}, MSMT17~\cite{msmt17}, DukeMTMC-reID~\cite{zheng2017unlabeledduke}, and Occluded-Duke~\cite{miao2019PGFAoccduke}.

\textbf{Market-1501} comprises a large-scale collection of images captured from a market setting. It contains 32,668 annotated bounding boxes of 1,501 identities, captured by six cameras. The dataset is divided into a training set with 12,936 images and a testing set with 19,732 images.

\textbf{MSMT17} is a large-scale person re-identification dataset collected from real-world scenarios. It consists of 126,441 images of 4,101 identities captured by 15 cameras. The dataset provides a training set with 32,621 images, a query set with 11,659 images, and a gallery set with 82,161 images.

\textbf{DukeMTMC-reID} is collected from the DukeMTMC project, which includes 8 high-resolution cameras covering different scenes. It comprises 36,411 images of 1,812 identities, divided into a training set with 16,522 images, a query set with 2,228 images, and a gallery set with 17,661 images.

\textbf{Occluded-Duke} is a variant of the DukeMTMC-reID dataset, which contains occlusion annotations. It includes 36,411 images of 1,812 identities, with occlusion annotations provided for 10,056 images. The dataset is divided into a training set with 16,522 images, a query set with 2,228 images, and a gallery set with 17,661 images.


We utilize the Cumulative Matching Characteristics (CMC) at Rank-1 (R1) and the mean average precision (mAP) as evaluation metrics to assess performance. 
Rank-1 (R1) refers to the accuracy metric in which the top-ranked candidate is considered correct. 
While mean Average Precision (mAP) is employed as a comprehensive evaluation metric that emphasizes the relative ranking and order of the retrieved results. 

\subsubsection{Implementation Details}

We utilize the image and text encoder with CLIP, and choose ResNet-50 and Vit-B-16 as the image encoder for the experiments, respectively. The maximum length of input sentences is set to 77. All person images are resized to 256 \(\times\) 128. The batch size is set to 64 with 4 images per ID. The training images are augmented with random horizontal flipping, padding, random cropping, and random erasing.  The image and text encoders are initialized with pretrained CLIP. The initial learning rate is set to \(5\times 10^{-6}\). For modules with random initialization, the initial learning rate is set to \(5\times 10^{-5}\).  We warm up the model for 10 epochs with a linearly growing learning rate from \(5\times 10^{-7}\) to \(5\times 10^{-6}\). The batch size is set to 64 with 4 images per identity. For the CNN-based model, We train the network for 120 epochs, then the learning rate is decayed by 0.1 at the 40th and 70th epochs. For the ViT-based model, We train the network for 60 epochs employing the Adam optimizer, which is decayed by a factor of 0.1 every 20 epochs. We extract the features from the lase layer of CFF module in CNN and VIT models, respectively, as the embeddings of each person image for ReID task.
All experiments are conducted on a single NVIDIA V100 GPU. 

\subsection{Comparison with State-of-the-arts}

Table~\ref{tab:performance} presents a comprehensive comparison of our proposed method with state-of-the-art approaches on various benchmark datasets including Market1501, MSMT17, DukeMTMC, and OccDuke. We evaluate the performance in terms of mean Average Precision (mAP) and Rank-1 (R1) accuracy. Our proposed method demonstrates superior performance compared to state-of-the-art approaches across all benchmark datasets, indicating its effectiveness in person re-identification tasks. Note that all data listed here are without re-ranking. As presented in Table 1, compared with the ViT-based method, the results of our method on the most challenging dataset MSMT17 achieves 88.2\% and 92.9\% on Rank-1 and mAP, respectively, surpassing previous state-of-the-art results. Especially, when compared with PromptSG which also leverages CLIP, we outperform it by 1.0\% and 0.3\% in Rank-1 and mAP, respectively. 
Our framework also demonstrates superior performance when adopting CNN as the backbone, largely outperforming its counterparts. 
Specially, Our method achieves 80.9\% mAP and 87.9\% R1, which are 12.4\% and 1.9\% higher than PromptSG. 
On other datasets, such as Market1501, DukeMTMC-reID, and Occluded-Duke, we achieve improvements in Rank-1 performance of 2.1\%, 1.3\%, and 0.4\%, respectively, when using the ViT-based backbone compared to CLIP-ReID. These results consistently demonstrate the superiority of CLIP-SCGI.

\subsection{Comparison of training efficiency}

As shown in Table~\ref{tab:training efficiency}, we compared CLIP-ReID and our proposed method, CLIP-SCGI, in terms of parameter count, training time, and accuracy on the Market-1501 and MSMT17 datasets. Despite CLIP-SCGI having a 26\% higher parameter count than CLIP-ReID, we observed no increase in training time for our one-stage approach compared to the two-stage CLIP-ReID.

Furthermore, our method demonstrates significant improvements in accuracy. CLIP-SCGI achieved mAP scores of 96.0\% on Market-1501 and 88.2\% on MSMT17, surpassing CLIP-ReID's scores of 89.6\% and 73.4\%, respectively. These results indicate that our one-stage training approach not only maintains training efficiency despite the increased parameter complexity but also achieves substantially higher accuracy compared to the conventional two-stage training of CLIP-ReID.

\begin{table}[tp]
    \centering
     \renewcommand{\arraystretch}{1.5} 
      \setlength{\tabcolsep}{7.5pt}
         \caption{Comparative Analysis of Parameter Count, Training Times, and mAP between CLIP-ReID and CLIP-SCGI (our method) on Market-1501 and MSMT17 datasets.}

    \begin{tabular}{c||c|c|c|c}
    \hline
          Dataset&Method&  \#Params&  Training Times& mAP\\
         \hline
          \multirow{2}{*}{Market-1501}&CLIP-ReID&  89M&  4689s& 89.6\\
          &CLIP-SCGI&  113M&  4451s&  96.0\\
          \hline
          \multirow{2}{*}{MSMT17}&CLIP-ReID&  90M&  12904s& 73.4\\
          &CLIP-SCGI&  113M&  13534s& 88.2\\
          \hline
    \end{tabular}
    \\[5pt]

    \label{tab:training efficiency}
\end{table}

\subsection{Ablation Studies}

We perform comprehensive ablation studies on MSMT17 and Market-1501 to scrutinize the impacts of major parts and all the experiments are conducted on ViT-B/16. 

\subsubsection{Ablation on the CGI module}
To gain deeper insights into the effectiveness of pseudo tokens \(S^*\) in providing fine-grained guidance for visual embeddings, we removed the CGI module and exclusively utilized ``a photo of a person'' as the input for the fusion module. Results presented in Table~\ref{ablition1}, No.2 vs. No.3 indicate a substantial impact of incorporating \(S^*\) on both mAP and R1 metrics. When \(S^*\) is included during training (No.3) model achieves higher performance compared to when \(S^*\) is excluded (No.2). Removal of \(S^*\)  during training resulted in a decrease in mAP of 2.4\% and 3\% on the Market1501 and MSMT17 datasets, and a decrease of 2\% on R1 for both datasets. These results demonstrate the importance of CGI in guiding the process of learning discriminative representations for ReID.

\begin{table}[tp]
    \centering
       \caption{Ablation of each component of
CLIP-SCGI on Market-1501 and MSMT17.}
    \renewcommand{\arraystretch}{1.5} 
    \setlength{\belowcaptionskip}{2mm}
    \setlength{\tabcolsep}{10pt}
    \begin{tabular}{c|c|c||c|c|c|c}
    \hline
        \multirow{2}{*}{No.} &  \multicolumn{2}{c||}{Components } & \multicolumn{2}{c|}{Market-1501}&\multicolumn{2}{c}{MSMT17} \\\cline{2-7}
     & CGI &  CFF  &  mAP&R1&mAP&R11 \\
       \hline
1& \checkmark & & 89.6& 95.5& 71.5&87.6\\
2&  & \checkmark & 93.6& 95.6& 85.2&90.9 \\
3&\checkmark &\checkmark  & 96.0& 97.6&88.2& 92.9 \\
  \hline

 \end{tabular}

    \label{ablition1}
\end{table}

\subsubsection{Ablation on the CFF module}
The effectiveness of the Contextual Feature Fusion module lies in its ability to facilitate better feature-level modal fusion. As shown in  Table~\ref{ablition1}, No.1 vs. No.3 compares the effectiveness of adding the Contextual Feature Fusion module. 
The first row represents a scenario where only the augmented training set is utilized, followed by supervised contrastive losses \(\mathcal L_{con}\) applied to the encoding of images and "a photo of a \(S^*\) person", without the addition of the fusion module. Compared to our full model, this approach results in a larger decrease in performance, further emphasizing the necessity of feature-level fusion. 
The CFF module effectively combines textual and image information, utilizing the semantic insights from textual descriptions and the visual features from images. This fusion results in a richer, more discriminative feature space, leading to improved accuracy and robustness in person re-identification. 

\subsubsection{Impact of semantic guidance in the CFF module}

In the feature fusion stage, we use semantic textual information "a photo of a \(S^*\)  person" as the query $Q$ for fusion. However, one may be curious: can one exploit the visual features? 
To verify its effectiveness, we further conduct experiments by replacing it with image features. Table~\ref{ablition2} reports the results. As shown, the replacement of the query with image features resulted in a decrease in mean Average Precision (mAP) of 1.9\%, 1.5\%, and 4.7 on the Market1501, MSMT17, and DukeMTMC datasets, respectively. Additionally, there was a reduction of 0.9\%, 0.4\% and 1.8 \%  in Rank-1 accuracy for both datasets.
These results indicate that changing the query from textual information to image features significantly impacts the performance of the person-identification system. 
This emphasizes the pivotal role of semantic guidance, where textual information directs the model's attention to relevant image regions, focusing on individuals rather than background elements. 
By leveraging this semantic guidance, our model achieves improved accuracy in re-identification tasks, as it gains a deeper understanding of the nuanced characteristics and attributes of individuals depicted in the image. This enables more effective matching and recognition of individuals across different images, ultimately enhancing the performance of our re-identification system, thereby facilitating more accurate person re-identification by minimizing the influence of irrelevant background elements.

\begin{table}[tp]
    \centering

     \caption{Ablation of replacing Q with image features in CFF.}
         \renewcommand{\arraystretch}{1.5} 
         \setlength{\belowcaptionskip}{2mm}
         \setlength{\tabcolsep}{9pt}
    \begin{tabular}{c||c|c|c|c|c|c}
    \hline
         \multirow{2}{*}{Method}&     \multicolumn{2}{c|}{Market-1501}&\multicolumn{2}{c|}{MSMT17} & \multicolumn{2}{c}{DukeMTMC}\\\cline{2-7}
         &  mAP&R1&mAP&R1 & mAP&R1 \\
       \hline
 & 96.0& 97.6& 88.2&92.9 & 81.5&90.6\\
 replace Q& 94.1& 96.7& 86.7& 92.5& 76.8&88.8\\
  \hline

 \end{tabular}
   
    \label{ablition2}
\end{table}

\subsubsection{Impact of CGI depth and number of learnable queries}
\begin{figure}[h]
 \setlength{\abovecaptionskip}{0.3cm} 
\centering
\subfloat[]{\includegraphics[width=0.49\linewidth]{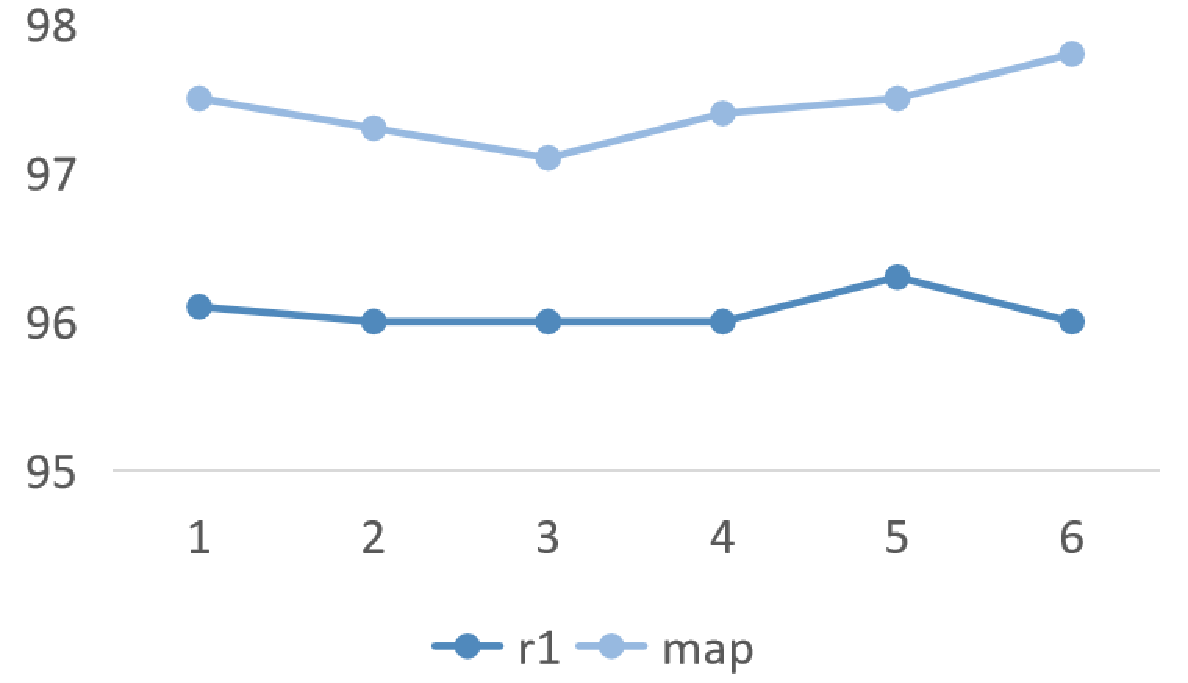}%
\label{fig:para1}}
\hfil
\subfloat[]{\includegraphics[width=0.49\linewidth]{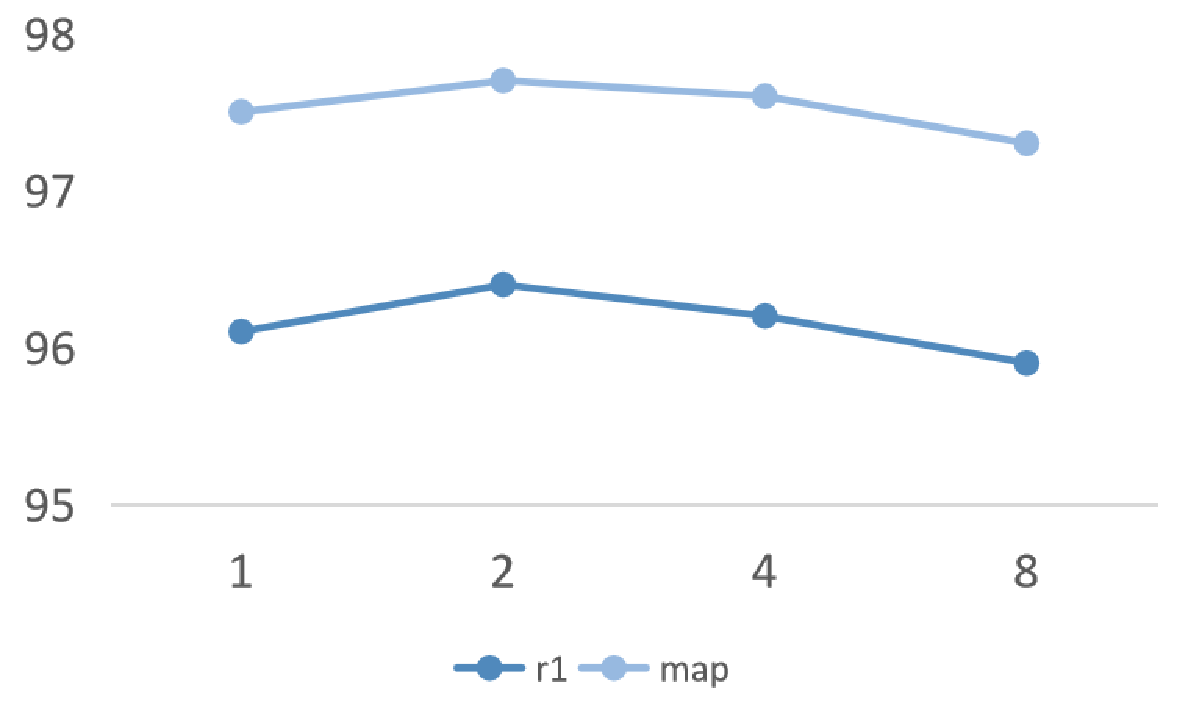}%
\label{fig:para2}}
\caption{The impact of the hyper-parameters at R1 and mAP on Market-1501, including (a) effects of CGI depth(left) ; (b) the number of learnable queries(right). The dark blue line represents R1, while the light blue line represents mAP.}
\label{fig-para}
\end{figure}

We evaluate the influence of varying the depth of the CGI  module. As shown in Fig.~\ref{fig-para}, our results reveal that a depth of 5 achieves the highest mAP, while a depth of 6 leads to the highest R1 performance. However, considering both performance and efficiency, we find that a depth of 1 strikes a favorable balance. We performed an additional ablation experiment on the length of the learnable prompt \(q_{K}\). As shown in Fig.~\ref{fig-para}\subref{fig:para2}, the range of \(q_{K}\) is {1, 2, 4, 8}. Setting the prompt length to 2 resulted in optimal performance.

\subsection{Visualization}
\subsubsection{Retrieval result visualization}
\begin{figure}[tp] 
\centerline{\includegraphics[width=\linewidth]{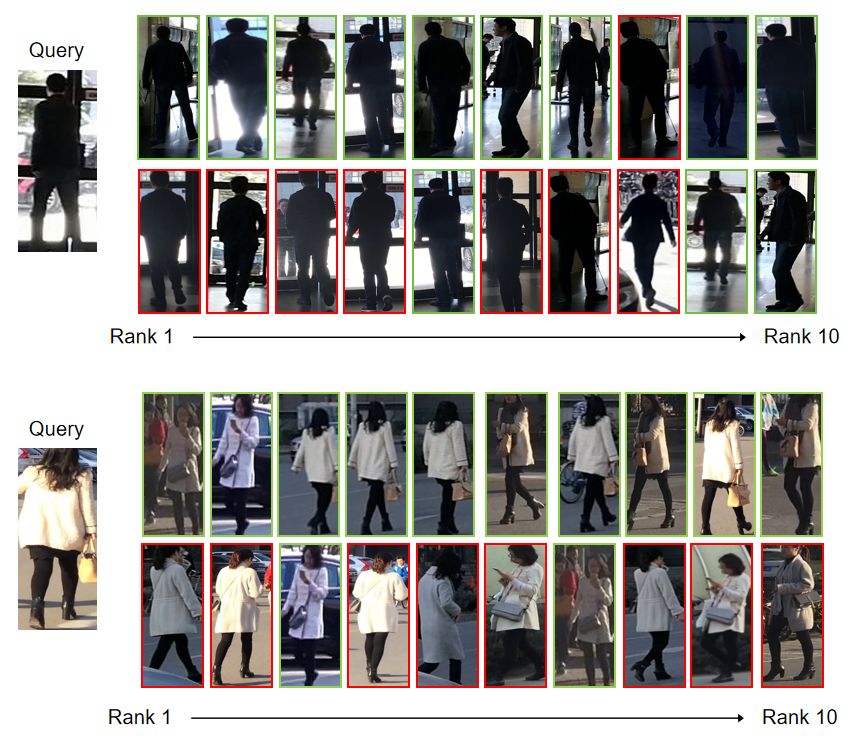}}
\caption{Comparison of top-10 retrieved results on MSMT17 dataset between our method (the first row) and CLIP-ReID (the second row) for each query.  The incorrectly identified samples are highlighted in red.}
\label{retrieval}
\end{figure}

Fig.~\ref{retrieval} exhibits two examples of top-10 retrieval results, where the first row and the second row present the results from our method and CLIP-ReID, respectively. Our method exhibits a notable enhancement as it excels in retrieving more challenging samples subsequently. The first example demonstrates a compelling contrast, indicating our method's robustness even under challenging conditions such as blurriness and low lighting. Remarkably, our approach manages to accurately identify subjects despite the adverse conditions, showcasing its capability for reliable recognition in real-world scenarios. Moreover, the second example in the results sheds light on another crucial aspect of our method's superiority. Even in scenarios where the person is similar, our method showcases its ability to discern subtle differences in features such as hair length and clothing, which are pivotal for accurate identification. This highlights the efficacy of our approach in capturing intricate details and leveraging them for precise recognition, surpassing the capabilities of existing methods like CLIP-ReID.

\subsubsection{Attention map visualization}
\begin{figure}[t]
\centering
\captionsetup[subfloat]{labelformat=empty} 
\subfloat[]{\includegraphics[width=9.5mm]{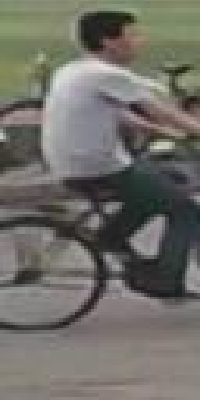}}
\hspace{0.1mm}
\subfloat[]{\includegraphics[width=9.5mm]{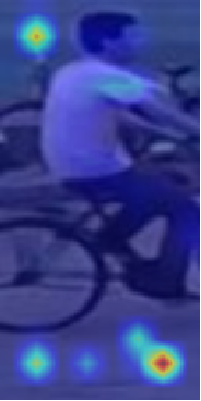}}
\hspace{0.1mm}
\subfloat[]{\includegraphics[width=9.5mm]{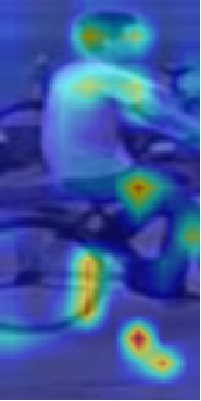}}
\hspace{0.1mm}
\subfloat[]{\includegraphics[width=9.5mm]{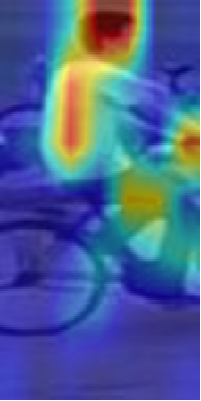}}
\hspace{2mm}
\subfloat[]{\includegraphics[width=9.5mm]{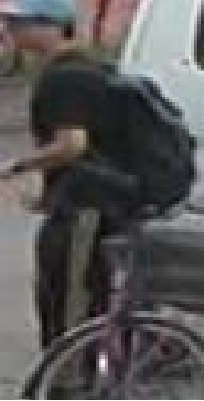}}
\hspace{0.1mm}
\subfloat[]{\includegraphics[width=9.5mm]{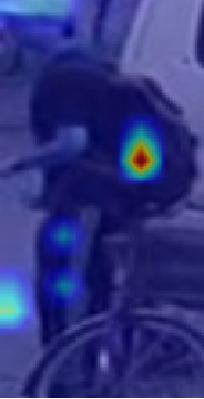}}
\hspace{0.1mm}
\subfloat[]{\includegraphics[width=9.5mm]{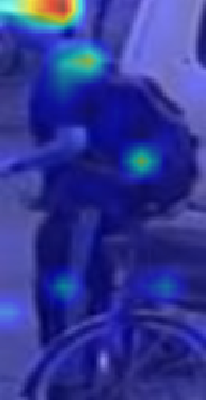}}
\hspace{0.1mm}
\subfloat[]{\includegraphics[width=9.5mm]{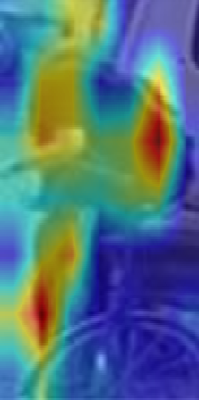}}
\captionsetup[subfloat]{labelformat=empty} 
\subfloat[(a)]{\includegraphics[width=9.5mm]{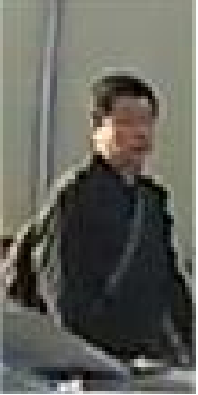}}
\hspace{0.1mm}
\subfloat[(b)]{\includegraphics[width=9.5mm]{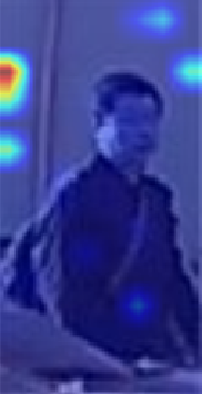}}
\hspace{0.1mm}
\subfloat[(c)]{\includegraphics[width=9.5mm]{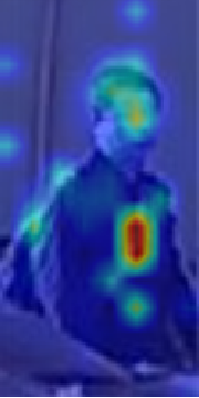}}
\hspace{0.1mm}
\subfloat[(d)]{\includegraphics[width=9.5mm]{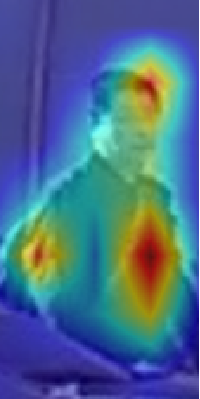}}
\hspace{2mm}
\subfloat[(a)]{\includegraphics[width=9.5mm]{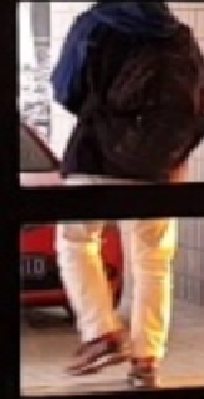}}
\hspace{0.1mm}
\subfloat[(b)]{\includegraphics[width=9.5mm]{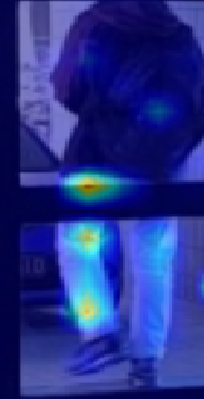}}
\hspace{0.1mm}
\subfloat[(c)]{\includegraphics[width=9.5mm]{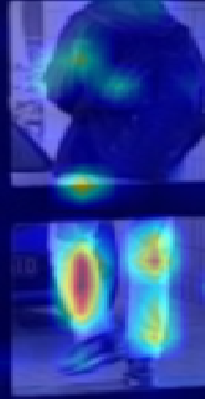}}
\hspace{0.1mm}
\subfloat[(d)]{\includegraphics[width=9.5mm]{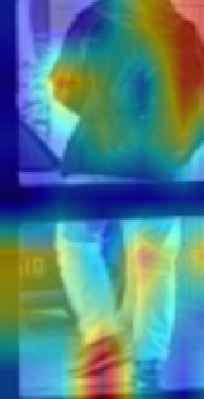}}
\caption{Attention map visualization on sample images from Market-1501 (top) and MSMT17 (bottom), respectively. (a) Input images, (b) baseline, (c) CLIP-ReID, (d) our CLIP-SCGI.}
\label{vis}
\end{figure}

To provide a more intuitive demonstration of the results, we conducted visualization experiments using the method~\cite{Chefer_2021_ICCV} to illustrate the model's focused areas. As is shown in Fig.~\ref{vis}, in the depicted images, the top row represents examples from the Market-1501 dataset, while the bottom row showcases instances from the MSMT17 dataset. It is evident that while the CLIP-ReID method can generally attend to the approximate locations of pedestrians, it falls short of capturing fine-grained details. Specifically, it struggles to identify crucial features of certain pedestrians, such as shoes, backpacks, and other distinguishing characteristics. These details are often pivotal for accurate identification, especially in challenging scenarios such as images with obscured or blurry backgrounds. In contrast, our approach, by incorporating semantic information, demonstrates superior capability in capturing detailed information and guiding the network to focus more precisely on pedestrians. This includes differentiating various body parts of pedestrians while disregarding the influences from the background, thus mitigating potential interference in the identification process.

\section{Discussion}

\begin{figure}[tp]
    \centering
 \setlength{\abovecaptionskip}{0.5cm} 
    \includegraphics[width=1\linewidth]{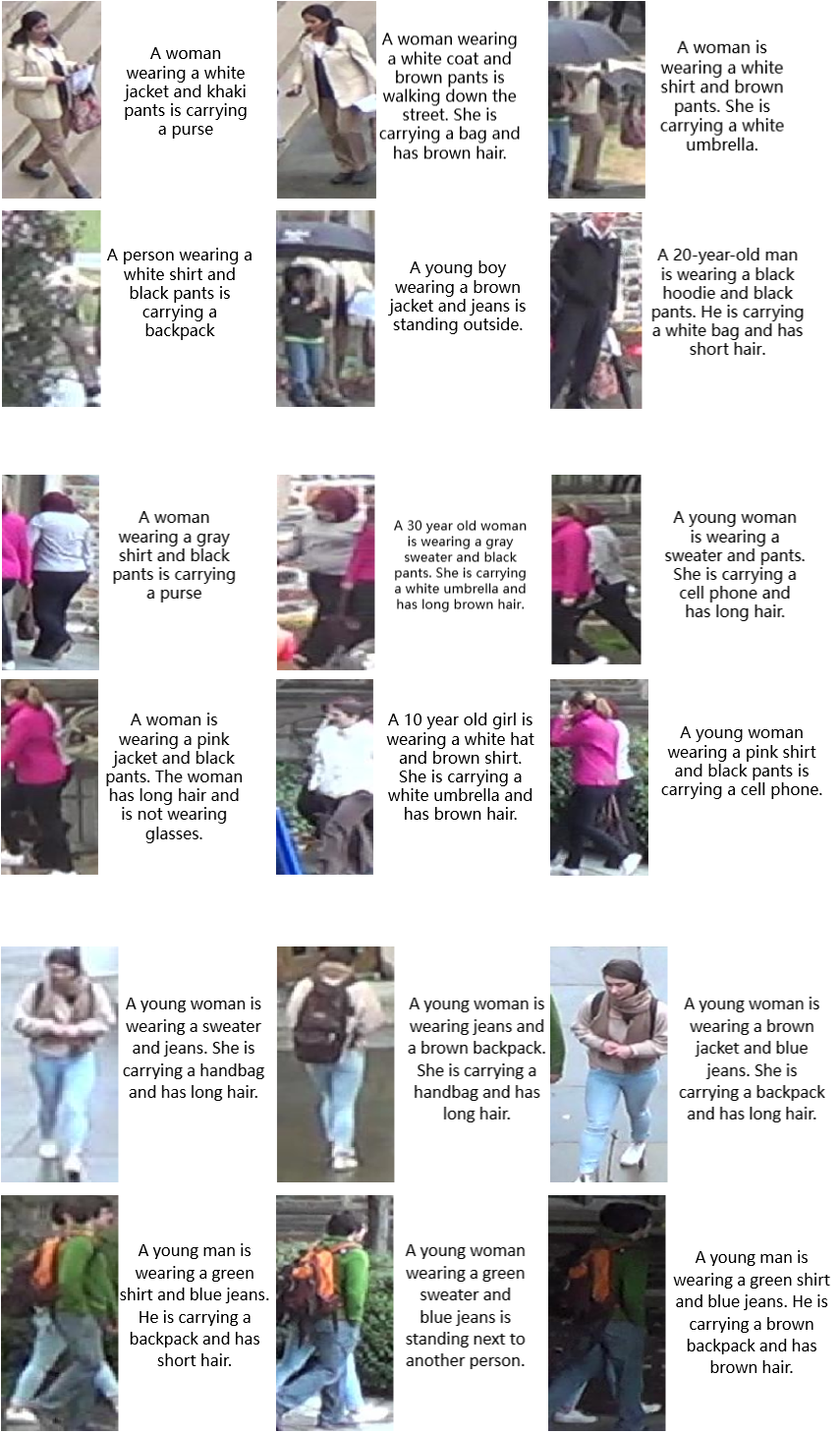}
    \caption{Caption Generation results on Occluded-Duke dataset. The figure illustrates examples of three pedestrians from the dataset. The first row showcases correct caption results, while the second row demonstrates captions generated under complete occlusion scenarios.}
    \label{fig:occ}

\end{figure}

\subsection{Limitations Due to Occlusion on Caption Quality }

As shown in Figure~\ref{fig:occ}, there are examples of three pedestrians in the dataset. The first row illustrates the captions generated under normal circumstances without occlusion. These captions can accurately capture the characteristics of the pedestrians in the picture, including gender, clothing, and other details. In contrast, the second row highlights the severe occlusion scenarios in the dataset where pedestrians are either nearly invisible due to intense obstructions or completely obscured by other individuals.

The limitations of image captions become noticeable here, as they struggle to accurately capture the facial characteristics or attributes of the obscured pedestrians. This can result in slightly inaccurate or less descriptive pseudo-words $S_*$ from the caption-guided inversion module. Caption inaccuracies can introduce subtle noise, potentially affecting the guidance for cross-modal fusion. As a result, the training process may encounter challenges in learning meaningful correlations between image and text embeddings, which could modestly impact the model's performance.

In summary, the accuracy of caption generation in occlusion scenarios is an area for improvement, ensuring better caption quality could further enhance the robustness of our model in addressing the complexities of occluded ReID tasks.

\section{Conclusions}

In conclusion, we present a novel approach to addressing the challenge of implicit text embeddings in person re-identification (Re-ID) by utilizing existing image captioning models to generate explicit descriptions for person images. By leveraging large vision-language models like LLAVA, we augment the traditional ReID training dataset from a unimodal (image) to a bimodal (text and image) framework, facilitating the efficient training of CLIP for ReID.
Our proposed CLIP-SCGI framework is both simple and effective, employing synthesized captions to guide the learning of ID-specific embeddings, thereby enhancing the extraction of discriminative and robust representations through language guidance. By embedding text and images into a joint space and incorporating a caption-guided inversion module, we enable the capture of corresponding semantic attributes within the visual images, leading to improved feature extraction and cross-modal fusion.
Extensive experiments on four widely recognized ReID benchmarks demonstrate the efficacy of our framework, revealing significant performance enhancements over state-of-the-art methods. Our approach not only streamlines the training process by providing concrete descriptions for person images but also achieves superior outcomes, underscoring its potential to advance the field of person re-identification.
However, our framework currently depends on offline image captioning models for generating high-quality captions. Future work will explore methods to eliminate this intermediary step by utilizing implicit text embeddings from large pre-trained multimodal models, thereby removing the necessity for explicit caption generation.

{
\bibliographystyle{IEEEtran}
\bibliography{reference}
}


\end{document}